\documentclass[11pt,a4paper]{article}
\usepackage[hyperref]{acl2020}
\usepackage{times}
\usepackage{latexsym}

\usepackage{graphicx}
\usepackage{microtype}
\usepackage{arydshln}
\usepackage{multirow}
\usepackage{placeins}
\usepackage{booktabs}
\usepackage{subcaption}
\usepackage{amsmath}

\usepackage{listings}
\usepackage{xcolor}

\usepackage{hyperref}

\makeatletter
\newcommand{\printfnsymbol}[1]{%
  \textsuperscript{\@fnsymbol{#1}}%
}
\makeatother

\aclfinalcopy 

\title{Cross-stitched Multi-modal Encoders}

\author{Karan Singla\thanks{equal contribution}, Daniel Pressel\printfnsymbol{1}, Ryan Price, Bhargav Srinivas Chinnari \\ {\bf Yeon-Jun Kim, Srinivas Bangalore}\\
Interactions Corp.}

\begin{document}
\maketitle

\begin{abstract}

In this paper, we propose a novel architecture for multi-modal speech and text input. We combine pretrained speech and text encoders using multi-headed cross-modal attention and jointly fine-tune on the target problem. The resultant architecture can be used for continuous token-level classification or utterance-level prediction acting on simultaneous text and speech. The resultant encoder efficiently captures both acoustic-prosodic and lexical information. We compare the benefits of multi-headed attention-based fusion for multi-modal utterance-level classification against a simple concatenation of pre-pooled, modality-specific representations. Our model architecture is compact, resource efficient, and can be trained on a single consumer GPU card.

\end{abstract}

\section{Introduction}

Speech interfaces have seen wide adoption through virtual assistants such as Siri and Alexa which have rapidly become a part of our everyday lives. To facilitate these applications, high quality automatic spoken-language understanding (SLU) components are essential. In a typical SLU, the Automatic Speech Recognition (ASR) system is used to convert speech into transcription hypotheses followed by a natural language understanding (NLU) component which acts on those hypotheses to extract an actionable semantic representation. However, in spoken language, organization of acoustic-prosodic cues within an utterance and in-between utterances can resolve semantic, lexical and syntactic ambiguities \cite{nagel1996prosodic, snedeker2003using, frazier2006prosodic}.

Most existing SLU systems first transcribe speech into text and then use the text as input to deep neural text encoders. It has been shown that providing speech-based features can aid text-based encoders, improving SLU for both chunk-level \cite{tran2017joint} and utterance-level applications \cite{chuang2004emotion, singla2018using}. However, most proposed chunk-level fusion methods generally use aligned speech and text to represent multi-modal chunk-level features \cite{tran2017joint}. Similarly for utterance-level fusion, most existing approaches do a simple concatenation of speech and text encoders before fine-tuning them for an SLU task. In this work, we propose to combine speech and text encoders using a jointly trained attention mechanism. As a result, every token in the text accounts for speech variability surrounding it without the need for an explicit alignment.

In this context, pretrained self-supervised encoders, which directly take the continuous input in the form of raw speech, have shown promising results when fine-tuned for transcription tasks. These encoders have also been successfully fine-tuned end-to-end for a variety of SLU tasks \cite{tzirakis2017end, chen2018spoken, ghannay2018end,yadav2020end}. Recently \cite{siriwardhana2020jointly} show that jointly fine-tuning pre-pooled speech and text encoders with simple concatenation can lead to improved results for emotion extraction. We start training from a pretrained Wav2vec2 model \cite{baevski2020wav2vec} for converting raw speech segments into fixed-dimensional temporal embeddings. In addition, we use a pretrained text encoder to convert text into token embeddings. We then apply a multi-headed attention between these embeddings in both directions, similar to encoder-decoder attention \cite{Bahdanau2015NeuralMT}.

The contributions of our paper is as follows:

\begin{itemize}
    \item We propose a cross-modal attention mechanism that does not require alignment between speech and text input.
    \item We apply the cross-modal representations to two token-level classification tasks (punctuation insertion in ASR hypothesis and speaker diarization based on ASR hypothesis) and show 2-4\% improvement over text-only model.
    \item  We also show improvements of 2-6\% over text-only models on intent and emotion identification -- both utterance-level classification tasks.
\end{itemize}

\section{Related work}
We briefly review  methods for learning text and speech-based self-supervised encoders. We then highlight a recent growing trend using pretrained speech encoders for high-quality SLU systems. Lastly, we briefly discuss the benefits of our proposed method in relation to previously proposed multi-modal SLU approaches.

\subsection{Self-supervised Representations}

Recently, it has become common practice to first pretrain text encoders using large amounts of unlabeled text before fine-tuning them for a target task \cite{Peters2017SemisupervisedST, Peters2018DeepCW, devlin2018bert}. A popular method of learning text-based, self-supervised encoders is to train a language model to predict the next word in a sequence \cite{mikolov2010recurrent, Radford2018ImprovingLU}. BERT \cite{devlin2018bert} introduced a Masked Language Model (MLM) objective, where tokens are randomly masked or perturbed and the model must learn to reconstruct those portions, yielding bidirectional representations.
This type of "self-supervision" has also been adopted to encode speech signals \cite{oord2018representation,pascual2019learning, chung2019unsupervised,baevski2019vq}. These encoders generally use training targets that are derived from the input signal. For example, the model may be tasked to recover the original input signal given a version transformed through augmentation techniques, recover masked inputs from the future or randomly in the sequence, or separate true inputs from synthetic samples. However, unlike text-based encoders, speech encoders generally need some amount of fine-tuning on a transcription task before being useful for SLU \cite{chorowski2015attention, chan2016listen, baevski2020wav2vec}.

\begin{figure*}[ht]
\centering
  \includegraphics[scale=0.65]{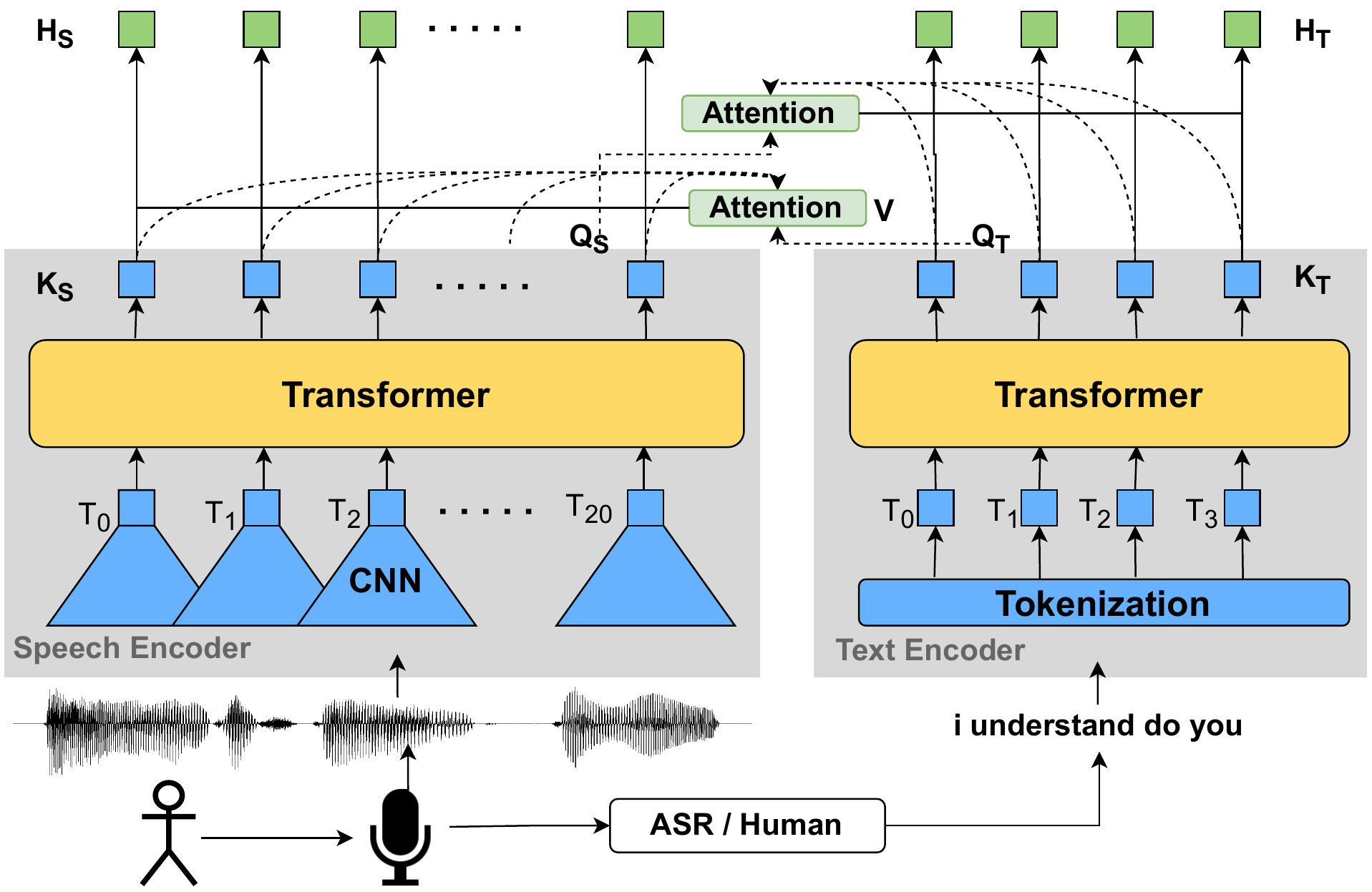}
  \caption{Cross-stitched encoding: Separately pretrained speech and text encoders are combined using a two-way multi-head cross-attention. Output of the attention-level gives token-level speech and text input which has attended to relevant information to decode a token by a supervised fine-tuning task. }
  \label{fig:speech_production}

\end{figure*}
\subsection{SLU directly from speech}
With the emergence of end-to-end ASR \cite{chorowski2015attention, chan2016listen} and the successful pretraining of speech encoders, methods for SLU directly from the speech signal have recently shown comparable performance to the conventional approach of cascading ASR and text-based components in tasks such as named entity recognition (NER), translation, dialogue act prediction (DAP) \cite{vila2018end, dang2020end}, as well as inference tasks like emotion, intent or behavior understanding \cite{fayek2015towards, price2020improved,singla2020towards}.

\subsection{Multi-modal SLU}

The speech features for multi-modal systems are generally provided either at the level of words or utterances based on the underlying SLU task. Combining speech and text features has led to improved results for  multiple tasks including: spoken text parsing, emotion extraction and also for automatic understanding of psychological disorders and human behavior \cite{yu2013multimodal,kim2019dnn,fraser2013using}.

The focus of this paper is to show benefits of cross-attention mechanism to perform fusion of continuous speech and text input stream, and an exploration of various self-supervised speech encoders \cite{arora2021espnet, hsu2021hubert, liu2021tera} was beyond the initial scope. Unlike previous multi-modal approaches \cite{kim2021st, you2021self}  it only needs aligned corpora for fine-tuning, not for pre-training. \cite{tsai2019multimodal} uses cross-attention between transformer layers of speech and text encoders for emotion extraction instead we use cross-attention on top of pre-trained off-the-shelf encoders. On MOSEI dataset they report 50.4\% vs 53.4\% accuracy for our system on 7-way sentiment prediction. We also plan to publicly release our models and setup.

In the past, similar token-level tagging approach has been proposed to spoken text parsing. They perform feature fusion of text and speech features, where speech features are simple functionals representing a word. sequence network for chunk-level multi-modal fusion methods \cite{tran2017joint}. Additionally, this is the first work, which performs multimodal token-level tagging to improve over speech only and text only approaches for diarization and rich transcription. Our proposed system can transparently perform multi-modal token-level tagging of speech and text without any supervised alignments. It learns to attend token from other modalities stream while performing tagging.

\section{Cross-stitched Multi-modal Encoder}

Cross-stitch\footnote{https://en.wikipedia.org/wiki/Cross-stitch} is a tiled, raster-like pattern $X$ used repeatedly to form a picture. We propose to combine pretrained speech encoder embeddings with temporal text encoder embeddings using two-way multi-headed cross-modal attention, which allows each encoder to attend to the other modality's encoder in every time-step. Figure 1 gives an overview of the architecture. Our pretrained speech encoder is first trained with Wav2vec2, then fine-tuned using transcribed data for an ASR task with a CTC loss. The text input is encoded using a pretrained MLM.

The speech and text encoders output $K_S$ and $K_T$ respectively. Keys $K_i$ are either text or speech tokens, and query $Q_j$ is output from the other modality.  Following the typical Transformer decoder approach, we first apply self-attention to the target query. Keys and queries are then connected using cross-attention similar to encoder-decoder multi-headed attention \cite{vaswani2017attention}. Queries and keys of dimension $[d_q, d_k]$, and values of dimension $d_v$ become inputs to the attention function. We compute the dot products of the query $Q_i$ with all keys $K_j$ and divide each by $\sqrt{d_{k}}$, where $d_{k_{j}}$ is dimensionality of keys we are attending. We then apply a softmax function to obtain the weights on the values.

\begin{align*}
{\displaystyle Attention(Q_i, K_j, V_j ) = softmax(\frac{Q_i*K_{S^j}}{\sqrt{d_{k_{j}}}})  * V_j}
\end{align*}

We then perform the attention operation $h$ times using different $V$ values where queries, keys and values are low-order projections using $W$, creating different representations at different positions in the other modality. We employ $h = 8$ parallel attention heads. 

Multihead cross-attention is formally defined as follows:

\begin{align}
    MultiHead(Q_i,K_j,V_j) = [head_1, ..,head_h]*W_j
\end{align}

where 

\begin{align}
head_n = Attn.(Q_n*W_Q^n, K_n*W_K^n, V_n*W_V^n)
\end{align}

where $W_i^n \in R^{d_{model}\times d_j}$  are parameter matrices. All heads {$[1:h]$} are concatenated to represent each multi-headed token-level cross-attention output for both speech and text input. An additional weight matrix $W_j$ then filters the information from these cross-stitched representations. We use the resultant multi-modal temporal output for various token-level tagging and utterance classification tasks described in later sections. 
All of our models and experiments are built with \cite{pressel2018baseline}, an open source library for model exploration and development targeting NLP\footnote{https://github.com/dpressel/mead-baseline}. 

\subsection{Speech Encoder}

For the speech encoder (SE), we use a Wav2vec2 model with 12 Transformer blocks with 12 attention heads and a 768 dimensional hidden unit size, similar to the base model in \cite{baevski2020wav2vec}. Our convolutional feature encoder is adapted for speech data sampled at 8kHz. The model was pretrained on approximately 9450 hours of anonymized speech data from a collection of conversational AI applications where users interact with an intelligent virtual agent (IVA) for customer care over the phone. The model was subsequently fine-tuned with a CTC loss on 900 hours of transcribed data \footnote{We saw consistent results with publicly available checkpoints.}. 

\begin{figure*}[ht]
  \centering
  \includegraphics[scale=0.65]{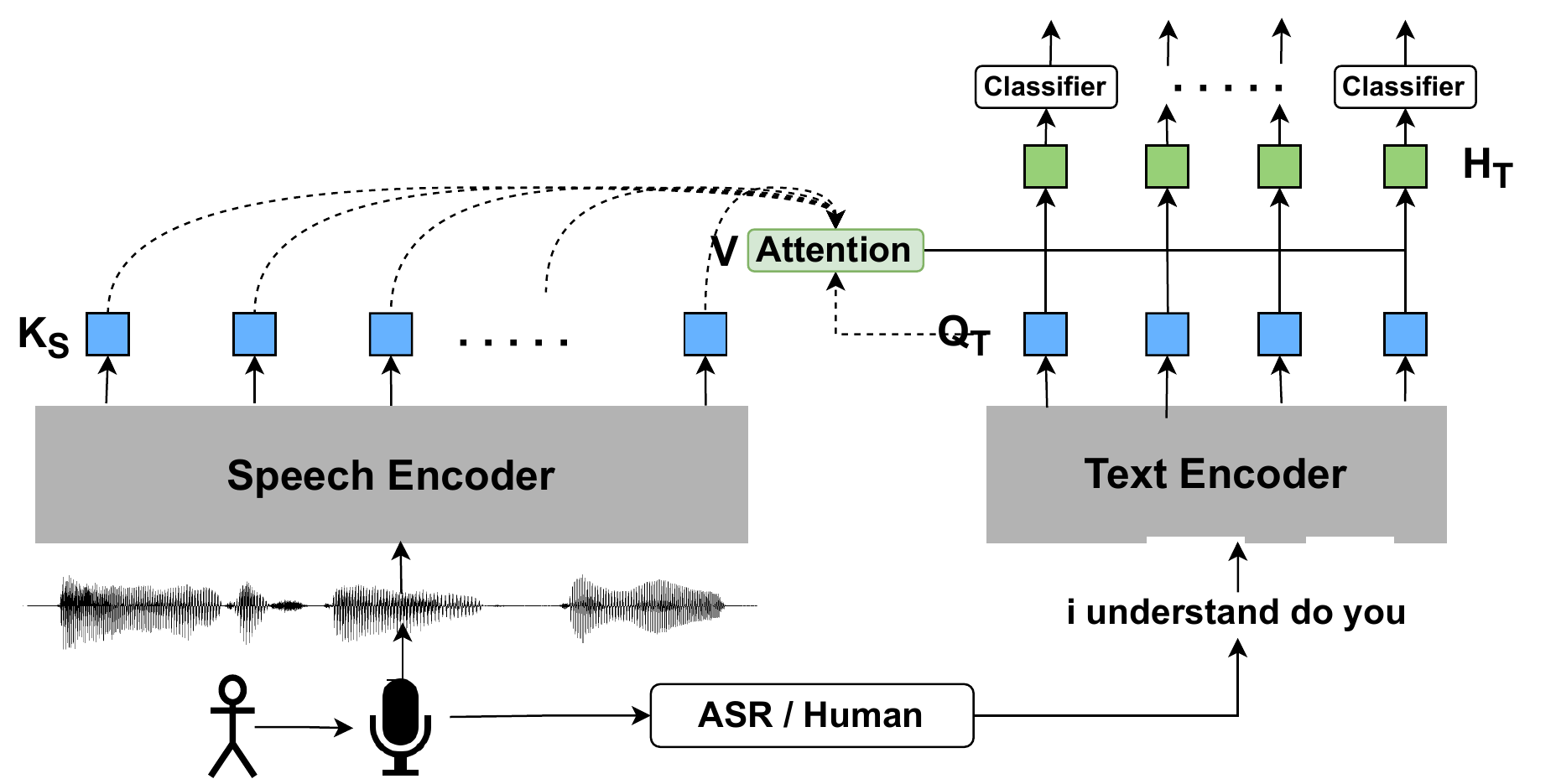}
  \caption{Word-level tagging using cross-attention mechanism. For each word-level prediction in text it takes cross-attn over the corresponding speech segment, thus, doing a soft alignment}
  \label{fig:speech_production_word}

\end{figure*}

Our initial testing showed that the lower layers of the architecture contributed most of the information relevant to downstream applications in the multi-modal setting. We found that removing the final 4 Transformer layers from the fine-tuned speech encoder resulted in very little change in performance, but significantly sped up training and inference, while reducing the overall memory footprint.  Subsequently, we dropped the final 4 layers of the speech encoder for all experiments.

\subsection{Text Encoder}

For the text-based encoder (TE), we pretrained an 8-layer Transformer, with 8 attention heads using an MLM loss on a corpus of online data including all of English Wikipedia, around 700 million conversations from Reddit \cite{AlRfou2016ConversationalCC, Henderson2019ARO}, 3.3 million online forums, and 8.2 million online reviews for restaurants and hotels.  The majority of the dataset contains full conversations between multiple users, and the turns are demarcated with a special end-of-utterance token. Following \cite{Shaw2018SelfAttentionWR}, we use relative positional representations which are not conditioned on the global position of the token but instead use a local relative offset embedding at every layer as part of the self-attention computation.  Previous literature has shown that placing the layer norm at the front of each sub-layer in the Transformer simplifies training and can improve performance \cite{Nguyen2019TransformersWT, Xiong2020OnLN, wang-etal-2019-learning-deep}, so we also follow this approach in our model.


We empirically observed in initial testing that the last 4 layers of the text encoder could be dropped in the downstream multi-modal application without significant performance degradation.  As a result, we truncate our text encoder to only the lower 4 of the original 8 layers.

\subsection{Training Details}

Our fine-tuning system is compact and lightweight and we are able to train with a single GPU -- even on a consumer card.  For most experiments, we use a single NVIDIA GTX 1080ti GPU.

We use Adam with a fixed batch size of 2 with a fixed learning rate of $1.0e-5$, for all experiments except for IVA intent detection, where we trained with a batch size of 16 on a single A100 GPU \footnote{We used a larger batch size due to the large size of the dataset, to compare against internal benchmarks, and because a grid search yielded significantly better results for that dataset.}. For all experiments, we keep the speech encoder frozen for the first 2000 steps of training. We calculate the cross-entropy loss of a final projection to the number of labels.  For tagging, this translates to token-level loss. We use early stopping on a validation set for all experiments.

\section{Token-level fine-tuning}

Our proposed cross-stitched network can be used for multi-modal token-level fine-tuning for both text and speech based classification tasks. In this paper, we focus on doing token-level classification of text tokens where it attends to temporal speech embeddings using multi-headed attention. Figure 2 portrays the multi-modal token-level tagging of text.

Rich transcription makes ASR results more readable and valuable for human users. We propose two rich transcription tasks as post-processing on ASR output: 1) Punctuation insertion \& capitalization and 2) Speaker diarization in role-based conversations.

\subsection{Punctuation insertion \& capitalization}

We gather data readily available data from Tatoeba\footnote{https://tatoeba.org/en/}, which provides sentences with punctuation and first-letter capitalization. It also includes speech for each sentence read by one or more speakers. In total we gather approximately 165K English sentences along with speech representing each sentence. We train our multi-modal system to insert punctuation, specifically, comma (Cm), period (Pr) \& question-mark (Qus) and also perform first-letter capitalization (Cp) of words. We use 141K, 12K and 13K samples for training, validation and testing respectively. We hypothesize speech has information which can help with punctuation insertion and word capitalization. In this work, our results are limited to the Tatoeba corpus. Training data is created by adding word-level tags for punctuation insertion and capitalization where input is the normalized text (see sample below).

\begin{table}[h]
\centering
\scalebox{0.75}{

\begin{tabular}{lllllll}
\textbf{Input}              & thank & you  & i    & understand  & do   & you   \\
\textbf{Word tags} & Cp:0  & 0:Pr & Cp:0 & 0:Pr        & Cp:0 & 0:Qus \\
\textbf{Output}    & Thank & you. & I    & understand. & Do   & you? 
\end{tabular}
}
\end{table}

Our system predicts 8 different tags (shown in Table \ref{table:puncrestore}) for each word input. Table \ref{table:puncrestore} shows word-level F1-scores for this task and illustrates the improvement in scores using the multi-modal approach ({\em XSE}) over text-only approach.

\begin{table}
\centering
\scalebox{0.97}{
\begin{tabular}{|cccc|}
\hline
\multicolumn{2}{|c|}{\textbf{Word-level tag}}                 & \multicolumn{2}{c|}{\textbf{ \% F1}}                \\ \cline{3-4} 
\multicolumn{2}{|l|}{}                                    & \multicolumn{1}{l}{Text} & \multicolumn{1}{l|}{XSE} \\ \hline

\emph{\underline{Punctuation}}           & \emph{\underline{Capitalization}}       & \textbf{}                 & \textbf{}                 \\
                             & {\color[HTML]{009901} Yes} &    83                       &         {\bf 87}                  \\
\multirow{-2}{*}{Comma (,)}  & {\color[HTML]{FE0000} No}  &        86                   &               {\bf 88}            \\
                             & {\color[HTML]{009901} Yes} &            97               &        {\bf 98}                   \\
\multirow{-2}{*}{Period (.)} & {\color[HTML]{FE0000} No}  &         100                  &                100           \\
                             & {\color[HTML]{009901} Yes} &             90              &            {\bf 94}               \\
\multirow{-2}{*}{Qus (?)}    & {\color[HTML]{FE0000} No}  &      99                     &           99                \\
None                         & {\color[HTML]{009901} Yes} &          100                &            100               \\
None                         & {\color[HTML]{FE0000} No}  &       100                    &             100              \\ 
{\bf Macro-average}                        &   &                93           &             {\bf 95}             \\ \hline
\end{tabular}
}
\caption{Results for Punctuation insertion and capitalization task comparing text-only vs proposed multimodal approach (XSE) on Toteba corpus.}
\label{table:puncrestore}
\end{table}

\subsection{Speaker diarization for role-based conversations}

Speaker diarization includes predicting speaker change and clustering segments to identify speakers. One widely adopted approach for unsupervised speaker diarization first segments the input speech into fixed-length frames using a fixed step size. These frame embeddings are then clustered for a session by performing hierarchical clustering using a pre-defined similarity measure. Supervised approaches are also used to learn speaker boundaries or perform end-to-end speaker diarization based on these speech frame embeddings. 

We cast speaker diarization as a token-level speaker tagging task. For this paper, we limit our study to conversations where speakers can have only two roles. We gather call-center conversation in the food domain between an agent and a customer. We gather human transcriptions, and annotations marking speaker boundaries and speaker roles for each segment. In all, we use 56 hours of speech for training, 10 hours for validation and 10 hours for testing purposes. Our evaluation set of 198 conversations contains 3.6K total speaker turns and 19K words which are tagged by our model to produce a diarizaed output. We hypothesize that because of assigned speaker roles there is a bias between speakers in terms of language use. Below is a sample encoding for two-person role-based conversations.

\begin{table}[h]
\centering
\scalebox{0.70}{

\begin{tabular}{ccccccccccc}
\multirow{2}{*}{\textbf{Mini-batch}}         & A0 & A1 & A2 & C0 & C1 & C2 & C3 & C4 & A0 & A1 \\
                                    & C0 & C1 & C2 & A0 & A1 & A2 & A3 & C0 & C1 & C2 \\
\multirow{2}{*}{\textbf{Word tags}} & 1  & 1  & 1  & 0  & 0  & 0  & 0  & 0  & 1  & 1  \\
                                    & 0  & 0  & 0  & 1  & 1  & 1  & 1  & 0  & 0  & 0 
\end{tabular}
}
\end{table}

Here Agent $(A0-A2)$ words are coded as 1 and client $(C0-C4)$ words as 0. We train the system to predict 0's and 1's in a continuous stream of words from ASR.

Speaker diarization performance is generally measured using Diarization Error Rate (DER), computed as a sum of false alarms (FA): silence being recognized as speech, missed detections (MD): speech being recognized as silence, and Speaker Error Rate (SER), the \% of incorrect speaker tags. In our speech-based results (upper part of Table \ref{table:speaker_diarization}\footnote{We ignore FA errors (at least 6\%) as they only account for silence regions in speech.}), we report error rates using a typical state-of-the-art speaker diarization approach. We first identify speech and non-speech regions using a Time Delay Neural Network (TDNN) classifier \cite{bai2019voice}. Each window of 1.5s length with an overlap of 0.5s is converted into 128-dimensional X-vector \cite{snyder2018x} by passing through an embedding network trained to classify the speakers of switchboard corpus \cite{godfrey1992switchboard}. We then measure similarity between x-vectors using Probabilistic Linear Discriminant Analysis (PLDA) \cite{ioffe2006probabilistic, prince2007probabilistic}. We found using additional unsupervised in-domain corpora (460 hours) translates to improved diarization performance. After measuring the similarity score between all  pairs of x-vectors using PLDA, they are clustered until we arrive at two clusters, one for each speaker in the recording. In our work, we found spectral clustering yields better performance than using standard Agglomerative Hierarchical Clustering (AHC) \cite{lin2019lstm}.

\begin{table}
\scalebox{0.80}{
\begin{tabular}{|cccc|}
\hline
\multicolumn{2}{|c|}{\multirow{2}{*}{\textbf{Approach}}}      & \multicolumn{2}{c|}{\textbf{\% Token error}}                     \\ \cline{3-4} 
\multicolumn{2}{|c|}{}                                        & \multicolumn{1}{c}{\textbf{SER+MD}} & \textbf{SER} \\ \hline

\multicolumn{3}{|l}{\emph{\underline{Speech time-series clustering}}}                                                          &        \\
\multicolumn{2}{|c}{VAD + Generic PLDA + AHC}            &          17.1                             & 14.4              \\
\multicolumn{2}{|c}{VAD + Generic PLDA + Spectral}       &                            10.7            &       7.7         \\
\multicolumn{2}{|c}{{\bf VAD + In-Domain PLDA + Spectral}}    & {\bf 7.4}                                   & {\bf 4.5}                \\ \hdashline
\multicolumn{2}{|c}{In-Domain PLDA + Spectral$*$} & 5.4                                   & 2.9                \\ \hline
\multicolumn{3}{|l}{\emph{\underline{Token-level role tagging}}}                                                         &        \\
Text (TE)                                &                    & \multicolumn{2}{c|}{8.1}                                   \\
{\bf XSE}                                     &                    & \multicolumn{2}{c|}{\textbf{7.6}}                                   \\ \hline
\end{tabular}
}
\caption{Token error rates for speaker diarization in 2-person call-center conversations. * is the result with speech vs non-speech segmentations provided by humans.}
\label{table:speaker_diarization}

\end{table}
The last two rows of Table \ref{table:speaker_diarization} shows results for our text-based speaker diarization approach using the cross-stitched encoder which improves over text based role tagging. For token-level word tagging based diarization, we treat word-level error as token error. Our cross-stitched multi-modal approach ({\em XSE}) shows improvements over text only baseline. Our text based diarization system shows similar performance when compared to a fully automated speech based unsupervised state-of-the-art approach without any in-domain unsupervised data. Best results are achieved for speech-based approach when human provided speech segment information is used instead of automatic voice activity detection (VAD) system.

\begin{figure}[h]

  \includegraphics[scale=0.49]{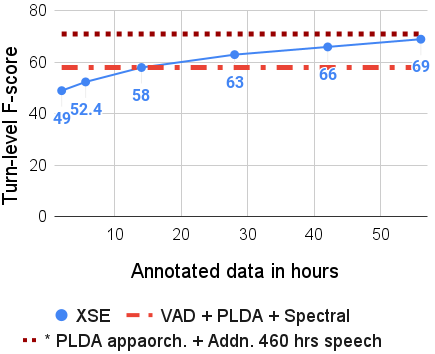}
  \caption{Comparison of multi-modal speaker diarization approach vs typical speech based diarization approach.}
  \label{fig:turnfscoresvsdata}

\end{figure}

Speech-based diarization performs global clustering of speech time frames versus token-level tagging of words which only uses local context. Therefore, we are unable to compare thse approaches directly at the token level. We propose a turn-level evaluation metric for two-person dialogues as high quality transcriptions also implies accurately the whole turn correct. We define Recall (R) as a ratio of number of correct turns to actual turns and Precision (P) is defined as the ratio of number of correct turns to detected turns. F-score is defined as $2PR/(P+R)$ irrespective of length of the segment. Figure \ref{fig:turnfscoresvsdata} shows variation of annotated data (speaker role and boundary information) along with turn-level diarization performance. Figure 3 shows results for multi-modal system using different sizes of annotated corpora. Our proposed approach performs similar to speech-based unsupervised PLDA approach with 14 hrs of annotated corpora. Text-only model shows 65\% turn-level F-score compared to 69\% for $XSE$.

Below is a sample output for our cross-stitched embedding (XSE) which takes normalized text as input. It shows combined output of punctuation insertion \& capitalization system and also diarization output by performing token-level role tagging.

\begin{table}[ht]
\centering
\scalebox{0.8}{
\begin{tabular}{|l|}
\hline
\textbf{Input}  \\
\begin{tabular}[c]{@{}l@{}}may i start with your phone number um five one\\ nine yes um let's see five one nine four two one \\ uh i don't phone myself so i don't know my damn\\ phone number um five three nine five three nine\\ four one two nine five three nine four one two\\ nine okay so is it pick up or delivery it's a delivery\end{tabular} \\ \hdashline
\textbf{Output} \\
\begin{tabular}[c]{@{}l@{}}A: May I start with your phone number?\\ C: Um five one nine.\\ A: Yes.\\ C: Um let's see five one nine four two one. Uh I \\      don't phone myself so I don't know my damn\\      phone number. Um five three nine five three\\      nine.\\ A: Four one two nine five three nine four one\\      two nine. Okay, so is it pick up or delivery?\\ C: It's a delivery.\end{tabular} \\ \hline
\end{tabular}
}
\end{table}

\begin{table*}[ht]
\centering
\scalebox{1.0}{
\begin{tabular}{|cccccccc|}
\hline
\multicolumn{8}{|c|}{\textbf{ Accuracy (\%)}}                                                                                                                                                                                                                                                                                                \\ \cline{1-8} 
\multicolumn{1}{|c}{\textbf{}}        & \multicolumn{1}{c}{\multirow{2}{*}{\textbf{Maj.}}} & \multicolumn{2}{c}{\textbf{Speech (SE)}}                                     & \multicolumn{1}{c}{\multirow{2}{*}{\textbf{Text (TE)}}} & \multicolumn{1}{c}{\multirow{2}{*}{\textbf{SE-TE}}} & \multicolumn{1}{c}{\multirow{2}{*}{\textbf{XSE}}} & \\ \cline{3-4}
\multicolumn{1}{|c}{\textbf{Dataset}} & \multicolumn{1}{c|}{}                                   & \multicolumn{1}{c}{\textbf{no-CTC}} & \multicolumn{1}{c|}{\textbf{with CTC}} & \multicolumn{1}{c}{}                                    & \multicolumn{1}{c}{}                                & \multicolumn{1}{c}{}                               &                                \\ \hline
MOSEI                                  & 40.7                                                    & 40.9                                 & 46.8                                   & 46.8                                                     & 51.7                                                   & {\bf 53.4}                                           &             \\
IVA                          & 57                                                      & 61.2                                 &                76.7                        & 79.5                                                     & 80.2                                          & {\bf 80.5}     &                                                                      \\ \hline
\end{tabular}
}

\caption{Results on emotion identification comparing our text-only approach against proposed multi-modal approaches.}
\label{table:emotion_eachdata}
\end{table*} 

\section{Utterance-level fine-tuning}

For spoken utterance classification we compare two fusion methods. First we adopt shallow fusion similar to \cite{siriwardhana2020jointly} by first pooling each individual encoder's output ($Q_S$) for speech and ($Q_T$) for text. The speech and text pooled output is then concatenated along the embedding dimension. For audio, we use max pooling, and for text, following BERT, we use the special start token ([CLS]). Some datasets contain samples with only text. For these samples, we sum along embedding dimension instead of concatenation to enable smooth training. {\em SE-TE} refers to shallow fusion and {\em XSE} refers to the cross-stitched encoder model in Table 5. The unimodal systems using pooling from either $Q_S$ or $Q_T$.

\subsection{Emotion Identification}



Creating a scalable general purpose solution for emotion extraction comes with the challenge of limited data annotations. \emph{Emotion} which captures behavioral information about a speaker has been primarily studied in the form of continuous or discrete perceived sentiment (negative, positive, neutral) \cite{zadeh2018multimodal, chen2020large}, 7 discrete emotions (anger, disgust, fear, joy, sadness, surprise) \cite{li2017dailydialog, busso2008iemocap} or more granular annotations of behavioral emotion \cite{demszky2020goemotions}.

In this paper, we study emotion as discrete annotations for spoken utterances which have both speech and text available. We use two different datasets that contain utterances labeled with discrete sentiment ranging from $-3$ to $3$. CMU-MOSEI \cite{zadeh2018multimodal} contains 23,453 annotated video segments from 1,000 distinct speakers and 250 topics, in total approximately 65 hours of speech along with transcriptions. Final sentiment annotated corpora contains 20k sentences annotated by 3 annotators. We follow the same data setup first as used by \cite{tsai2019multimodal}. We also use spoken utterances marked with discrete 7-way sentiment annotated data from an Intelligent Virtual Assistant (IVA) system in the customer care domain. We collect 10K unstructured spoken customer utterances from human-machine dialogue. These utterances/sentences are then coded for sentiment by 3 human annotators, with an agreement of about 75\%. We use 8K for training, 1K for development and 1K for testing purposes. We mix data from all annotators for train and test. Neutral (0) is the dominating label in both datasets, which is also the majority class performance shown in Table \ref{table:emotion_eachdata}. Our fusion approaches shallow fusion ($SE-TE$) and cross-stitched fusion ($XSE$) both outperform text only baselines. $XSE$ performs better than $SE-TE$ for both the MOSEI and IVA dataset. Our shallow fusion system $SE-TE$ is similar to \cite{siriwardhana2020jointly} as both concatenate the pooled encoder outputs before classification, however, we use a conversationally-trained, compact MLM instead of the original BERT encoder.

\subsection{Intent Detection}

\begin{table*}[ht]
\centering

\begin{tabular}{|ccccc|}
\hline
\textbf{Dataset} & \textbf{Speech (SE)} & \textbf{Text (TE)} & \textbf{SE-TE} & \textbf{XSE} \\
\hline
IVA & 82.34\% & 83.07\% & 84.01\% & \textbf{84.23\%} \\
FSC & 99.58\% & 99.34\% & 99.53\% & \textbf{99.63\%} \\

\hline
\end{tabular}

\caption{Intent detection on IVA and FSC dataset with different modalities}
\label{table:intent_detection}
\end{table*}

Intent detection -- attempting to understand a user's goal in a task-oriented dialogue -- is a typical problem in SLU.  It has primarily been treated as an unstructured prediction problem, applied either independently, or jointly with a separate task to collect specific named entities specific to a conversation (also referred to as slot-filling).  For text-only systems, the input to an intent-detection classifier would commonly be text composed by the user, but for spoken systems, the ASR pipeline is typically run first, yielding either a speech lattice or, more commonly, a list of the top transcription hypotheses from the ASR system (referred to as N-best lists).

{\bf Intelligent Virtual Assistant} We use a large dataset collected from a real-world virtual assistant applications in the customer care domain. It contains approximately 1.1 million anonymized utterances for training. Due to the size of the training set and the cost associated with obtaining human transcription of the spoken utterances and intent labels, N-best hypotheses for the spoken text are taken from a production ASR system consisting of a hybrid DNN-HMM acoustic model and an N-gram language model. Word accuracy of this ASR system is estimated to be in the mid to upper 80\% range for this data. The intent labels for training come from two sources. The labels are either generated automatically by an existing production SLU system when the confidence of the system is very high, or the utterances are sent to a human agent in-the-loop to be manually labeled when the confidence of the automated label is low. The test set consists of approximately 11K utterances that are manually labeled and verified.  A development set of approximately 38K noisily annotated utterances is used for early stopping. The dataset has 2 sets of labels indicating intent and entity predictions and, for classification, we use a multi-headed classifier to predict both.  The joint accuracy is used to indicate overall performance.  For the text modality, the N-best hypotheses are concatenated using a special end of utterance demarcation token (the same end-of-utterance token seen in text pre-training) and passed into the text encoder.

{\bf Fluent Speech Commands:} We use the publicly available Fluent Speech Commands (FSC) dataset \cite{lugosch2019speech} to train and evaluate our model and compare with models tested on the same dataset. The FSC corpus is the largest freely available spoken language understanding dataset that has intent labels using a wide range of subjects to record the utterances. In total, there are 248 different distinct phrases in the FSC dataset
and 5 distinct domains. The data are split into 23,132 training
samples from 77 speakers, 3,118 validation samples from 10 speakers
and 3,793 test samples from 10 speakers. Using human transcriptions our text encoder alone can achieve 100\% accuracy. However automatically generated transcripts using ASR are generally noisy. We use the two most likely transcripts generated using an end-to-end ASR model trained with NeMo toolkit \footnote{https://catalog.ngc.nvidia.com/orgs/nvidia/collections/nemo\_asr}. We then use these transcriptions as input to our text encoder.

For the FSC dataset, we observe that, while simple concatenation of the embeddings does not outperform the audio-only encoder, our cross-attention method does better despite a much lower accuracy for the text-only modality (Table \ref{table:intent_detection}).

\section{Conclusion}
Our results show that cross-stitching speech and text encoders using multi-headed attention produces strong results on a diverse set of datasets. Our proposed method supports continuous multi-modal tagging for speech and text input streams. We believe our results can be improved further by including task specific data into unsupervised pretraining of speech and text encoders and exploiting context in dialogue for utterance classification. We plan to explore these directions and evaluate our approach on additional tasks in the future.

We believe our system can be made more robust for near real-time streaming by training with longer sequence lengths and/or by exploiting the context. We plan to extend our approach to more tasks including inverse text normalization, named entity recognition and sentiment tree parsing. We believe our cross-stitching approach can be extended to other modalities e.g: using an additional visual encoder.

\section*{Acknowledgements}

Text-based MLM pre-training supported with Cloud TPUs from Google's TPU Research Cloud (TRC)

\bibliography{anthology,acl2020}
\bibliographystyle{acl_natbib}

\end{document}